\documentclass[10pt,letter,onecolumn]{article}
\usepackage[margin=1.0in]{geometry}
\usepackage{url}
\usepackage{hyperref}
\usepackage{graphicx}
\usepackage{multirow}
\usepackage{algorithm}
\usepackage{algorithmicx}
\usepackage{algpseudocode}
\usepackage{fancyvrb}
\usepackage{caption}
\usepackage{subcaption}
\usepackage{wrapfig}
\usepackage{capt-of}

\makeatletter
\renewcommand{\ALG@beginalgorithmic}{\scriptsize}
\makeatother

\title{
    \textbf{Pathfinding in Random Partially Observable
    Environments with Vision-Informed Deep Reinforcement
    Learning}
}

\author{Anthony Dowling}

\date{
    \vspace{-0.60em}
    \small Clarkson University \\
    \small Potsdam, NY 13699, USA \\
    \small dowlinah@clarkson.edu
}

\begin{document}

\maketitle
%\thispagestyle{empty}
%\pagestyle{empty}

%\vspace{-2.5em}

\begin{abstract} 
    
    Deep reinforcement learning is a technique for solving
    problems in a variety of environments, ranging from
    Atari video games to stock trading. This method
    leverages deep neural network models to make decisions
    based on observations of a given environment with the
    goal of maximizing a reward function that can
    incorporate cost and rewards for reaching goals.  With
    the aim of pathfinding, reward conditions can include
    reaching a specified target area along with costs for
    movement. In this work, multiple Deep Q-Network (DQN)
    agents are trained to operate in a partially observable
    environment with the goal of reaching a target zone in
    minimal travel time. The agent operates based on a
    visual representation of its surroundings, and thus has
    a restricted capability to observe the environment. A
    comparison between DQN, DQN-GRU, and DQN-LSTM is
    performed to examine each models capabilities with two
    different types of input. Through this evaluation, it is
    been shown that with equivalent training and analogous
    model architectures, a DQN model is able to outperform
    its recurrent counterparts.

\end{abstract}

%KEYWORDS
{\small 


%\textit{\textbf{Note:}} This research document is referred to as ``\textit{this work}'' throughout
%the document.

}

\section{Introduction}
\label{sec:intro}

Deep Reinforcement Learning (DRL) has been applied to
pathfinding in environments ranging from robot and drone
operation~\cite{robot2020, primal2019} to maritime
navigation~\cite{ships2019}. This technique has shown strong
capabilities to create models that make intelligent
decisions in these environments~\cite{ships2019}.  The DRL
models are used to control entities acting in a given
environment.  These entities are referred to as
``agents''~\cite{robot2020}.
%These models control entities that act in an environment
%are referred to as ``agents.'' 
Agents can utilize varying underlying deep learning models
to make decisions, such as Multi-Layer Perceptron
(MLP)~\cite{tracking2021}, Recurrent Neural Networks
(RNN)~\cite{robot2020}, Convolutional Neural Networks
(CNN)~\cite{grid2018}, and others.

Optimal choice of the model depends on the information
available in the target environment. For instance, in a
fully observable environment, like a board game, a model
with a memory capability is likely not
necessary~\cite{tracking2021}.  However, an agent operating
in a partially observable environment can benefit from a
model that has a recall capability, such as an
RNN~\cite{robot2020}. This class of environment can include
scenarios such as autonomous vehicles and drones, certain
classes of card games, and other instances where not all
information pertinent to decision making is available at a
given instant.  In such environments, the decision making of
a memory-capable neural network can benefit from past
information~\cite{survey2021}.

Many works apply Reinforcement Learning (RL) to the problem
of pathfinding in a variety of environments, where some of
which are partially
observable~\cite{robot2020,primal2019,ships2019}.  There are
many instances where pathfinding must be achieved without
complete knowledge of the environment. A prime example of
this is vision-based navigation. In this case, an agent is
set to search for a target in the environment based on
visual input~\cite{longrange2018}. This leads to an
inherently restricted observation space, as the agent is
only able to see a portion of the environment from a given
position. For instance, the agent will not be able to see a
target that is behind it. In other instances, there can be
obstacles included in the environment that may completely or
partially occlude the agent's view of the target.

Many works utilize a gridworld environment, where the agent
learns to navigate an environment that is represented as a
matrix of cells~\cite{primal2019,grid2018,restricted2021}.
This enables path finding in realistic environments, but
requires some overhead to translate the real environment
into a grid world to make use of the model. Instead, by
creating a training environment where the agent is informed
based on depth vision, the trained agent is more ready to
immediately operate in a realistic environment without the
added translation into a grid world.

In this work, the modelled environment is treated as 2
dimensional and is partially observable to the agent.  The
agent's vision consists of a label for the object in view,
and the distance to it.  This method can be readily expanded
for application in a 3 dimensional environment, but by
applying restricting the environment to 2 dimensions, the
computational time needed to investigate this pathfinding
problem are lowered to a more tractable level. The major
contributions of this work are as follows:

\begin{itemize}
    \itemsep0em
    \item Design and implementation of an RL environment
        that expands the grid world environment concept for
        training models informed by depth-based visual
        information
    \item Design of a novel reward function to encourage
        effective training of RL models for pathfinding
    \item Training of RL models on randomized environment
        maps that are shown to be solvable with BFS and
        rising environment difficulty through a variable
        number of obstacles
    \item Performance comparison of three RL models during
        training and evaluation
    \item Examination of model performance when knowledge of
        the target location is known versus unknown
\end{itemize}

The remainder of this work is structured as follows:
Section~\ref{sec:related} contains a survey of related
works, Section~\ref{sec:env} describes the methods for
environment generation for model training and evaluation.
Section~\ref{sec:rl} describes the reinforcement learning
configuration. Section~\ref{sec:eval} describes the model
evaluation. Lastly, Section~\ref{sec:conc} concludes this
work with a discussion of future expansion.

\section{Background and Related Work}
\label{sec:related}

Panov et al. apply RL to the pathfinding problem using
Q-learning~\cite{grid2018}. In their work, the environment
map is represented by a cell matrix, where each cell can be
either traversible or have an obstacle. The agent is able to
view a 5$\times5$ section of the map. The agent is placed at
the center of the 5$\times$5 portion. The work shows that
the Q-learning method using a CNN for feature extraction is
able to find valid paths in their partially observable
environment~\cite{grid2018}.

In a work similar to Panov et al.~\cite{grid2018}, Pe\~{n}a
and Banuti aim to evaluate reinforcement learning for
pathfinding in environments with varying
complexities~\cite{restricted2021}. Pe\~{n}a and Banuti use
a modified version of the Open AI Frozen Lake environment,
which is a 2D pathfinding environment that includes hazards
to the agent, along with ``slippery'' tiles that introduce a
random effect to agent actions within the environment. This
work utilizes a limited observation space to examine the
generalization ability of a Deep Q-Network (DQN) model. This
is acheived by training the agent on a set of three maps,
and evaluating its performance using unseen
maps~\cite{restricted2021}. In contrast, this work utilizes
camera-like vision to inform the model of the environment,
and does not include random effects on the agent.
Additionally, the training maps of this work are randomized
so that the agent does not train on the same map twice.

Sartoretti et al. further expand the pathfinding problem to
involve multiple agents~\cite{primal2019}. Their work uses a
gridworld in which the view is centered on each agent,
similar to the previous two works. To avoid collisions
between agents, the model is trained using a reward function
that also penalizes collision heavily.  Furthermore, selfish
behavior is avoided by using a combination of RL and
Imitation Learning to allow for an expert function to
penalize agents for blocking the movement of other
agents~\cite{primal2019}. In contrast, this work does not
investigate the multi-agent scenario.

Chen et al. implement a knowledge-free Q-learning model to
plan paths for smart ships~\cite{ships2019}. Faust et al.
implement RL for probabilistic roadmaps, which aims to aid
in long-range robot navigation using RL and sampling-based
path planning~\cite{longrange2018}.  Quan et al. use a
recurrent double DQN architecture for robot path-planning in
a gridworld environment~\cite{robot2020}.  Lastly, Sugimoto
et al. examine the tracking capabilities of a DQN
model~\cite{tracking2021}. These studies show applications
of reinforcement learning to the path planning problem, but
there are relatively few that utilize vision as input to the
model, or that compare DQN models with recurrent versions as
is done in this work.

\section{The Seeker Environment}
\label{sec:env}

\begin{algorithm}[h]
    \caption{Gridworld Environment Generation}
    \label{alg:envgen}
    \begin{algorithmic}[1]
        \Function{GetEmptyCell}{$cells$}
            \State $x \gets Random \: integer \: s.t. \: 0 \leq x < w$
            \State $y \gets Random \: integer \: s.t. \: 0 \leq y < h$
            \While{$cells[y,x] \neq 0$}
                \Comment{Regenerate coordinates to get blank location as needed}
                \State $x \gets Random \: integer \: s.t. \: 0 \leq x < w$
                \State $y \gets Random \: integer \: s.t. \: 0 \leq y < h$
            \EndWhile
            \State \textbf{return} $[x,y]$
        \EndFunction
        \Function{EnvGeneration}{$w,h,o$}%\Comment{$m$ and $n$ are the number of rows and columns in the matrix environment}
            \State $cells \gets h \times w \:\:zeroes$
            \For{$i\gets 0:o$}\Comment{Generate Obstacle Locations}
                \State $x,y \gets GetEmptyCell(cells)$
                \State $cells[y,x] \gets 1$\Comment{Store obstacle location in matrix}
            \EndFor
            \State $x,y \gets GetEmptyCell(cells)$
            \State $cells[y,x] \gets 2$\Comment{Store target location in matrix}
            \State $x,y \gets GetEmptyCell(cells)$
            \State $cells[y,x] \gets 3$\Comment{Store agent location in matrix}
            \State \textbf{return} $cells$
        \EndFunction
    \end{algorithmic}

\end{algorithm}

Many works utilize gridworld environments to evaluate RL
pathfinding models. These environments are useful, but they
are a restrictive abstraction of a realistic environment. To
more closely model a real-world environment, in this work, a
solvable gridworld map is generated. The solvable gridworld
map is used to randomly generate the positions of the
entities (agent, obstacles, and target). This map is then
converted into a 2-dimensional plane, allowing for precise
Euclidean distance measurements between two entities (any
combination of agent, obstacles, and target). Instead of an
obstacle being a cell in a matrix that the agent cannot move
across, obstacles become 4-sided walls that block the
forward movement of the agent.  This forces the agent to
operate by either turning or moving forward, allowing the
output of the RL model to directly match actions that would
be performed in a real environment.  This reinforcement
learning environment has been dubbed the ``Seeker''
environment.

\begin{figure}
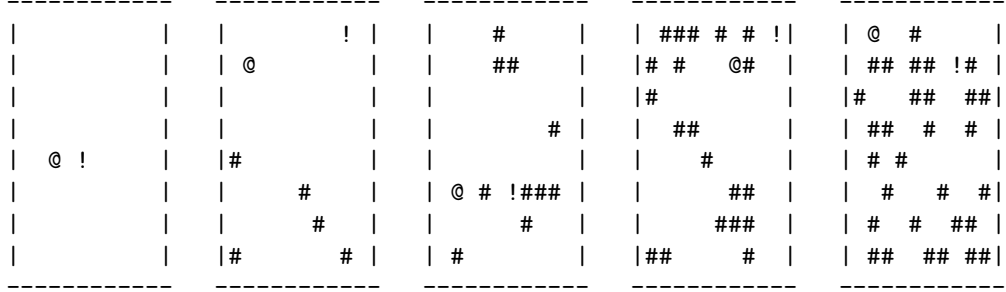

    \centering
    \begin{BVerbatim}
------------   ------------   ------------   ------------   ------------
|          |   |        ! |   |    #     |   | ### # # !|   | @  #     |   
|          |   | @        |   |    ##    |   |# #   @#  |   | ## ## !# |   
|          |   |          |   |          |   |#         |   |#   ##  ##|   
|          |   |          |   |        # |   |  ##      |   | ##  #  # |   
|  @ !     |   |#         |   |          |   |    #     |   | # #      |   
|          |   |     #    |   | @ # !### |   |      ##  |   |  #   #  #|   
|          |   |      #   |   |      #   |   |     ###  |   | #  #  ## |   
|          |   |#       # |   | #        |   |##     #  |   | ##  ## ##|   
------------   ------------   ------------   ------------   ------------
    \end{BVerbatim}
    \caption{Example Random Gridworld Environments}
    \label{fig:gridworld}
\end{figure}

To generate the gridworld, the \texttt{EnvGeneration}
function from Algorithm~\ref{alg:envgen} is used. This
function takes the gridworld width, height, and the number
of obstacles as input. Then, the \texttt{GetEmptyCell}
function is used to find available positions via brute force
to insert obstacles (Line 12-14) before inserting the agent (Line 19) and the
target (Line 17) into their respective randomly selected cells. This
yields environments such as those shown in
Figure~\ref{fig:gridworld}. The configured width of the
shown maps is 10, and the height is 8. The ``\texttt{-}''
and ``\texttt{|}'' symbols represent the edges of the map,
``\texttt{@}'' is the agent, while ``\texttt{!}'' is the
goal, and ``\texttt{\#}'' represents an obstacle.  The
number of obstacles in the maps shown in
Figure~\ref{fig:gridworld} varies from 0 to 30. 

%BFS for reachability checking.
\begin{algorithm}[h]
    \caption{BFS for Verifying Goal Reachability}
    \label{alg:bfs}
    \begin{algorithmic}[1]
        \Function{CheckReacability}{$cells$,$x_{agent}$,$y_{agent}$,$x_{target}$,$y_{target}$}
            \State $w,h \gets width(cells), height(cells)$
            \State $visited \gets [ [ x_{agent}, y_{agent} ] ]$
            \State $moves \gets [ [-1,0],[0,1],[1,0],[0,-1] ]$
            \Comment{Left, Up, Right, Down $(x,y)$ move pairs}
            \While{1}
                \State $visited.append([])$
                \For{$coor \in visited[-2]$}
                    \Comment{For every cell visited in the last iteration,}
                    \For{$move \in moves$}
                        \Comment{For every possible move from that cell}
                        \State $n \gets coor+move$
                        \If{$n_x>w$ or $n_y>h$ or $n_x<0$ or $n_y<0$}
                            \Comment{Check that $n$ is within the gridworld}
                            \State continue
                        \EndIf
                        \State $c \gets cells[n_y][n_x]$
                        \If{$c == $``!''}
                            \Comment{Target Found}
                            \State \textbf{return} True
                        \ElsIf{$c ==$``~''}
                            \Comment{Empty, Unvisited Cell Found}
                            \State append $n$ to $visited[-1]$
                            \State $cells[n_y][n_x] ==$``.''
                        \EndIf
                    \EndFor
                \EndFor
                \If{$length(visited[-1]) == 0$}
                    \Comment{Finding no empty cells in an iteration terminates the search}
                    \State \textbf{return} False
                \EndIf

            \EndWhile
        \EndFunction
    \end{algorithmic}

\end{algorithm}

Algorithm~\ref{alg:bfs} demonstrates how Seeker uses
Breadth-First Search (BFS) to check for goal reachability.
Starting at the agent, all moves from the agent one cell in
each cardinal direction (up, down, left, and right) are
examined.  If the target is found, ``true'' is returned. For
any cell neighboring the agent, if they are empty, its
position is noted in the $visited$ array, and its position
in the $cells$ matrix is marked. Then, the sequence repeats,
checking valid moves from the cells that were visited in the
last iteration. This algorithm fills the cells matrix with
``.'' until it either finds the target or runs out of valid
moves.  Running out of valid cells to move to in an
iteration indicates that the target is unreachable from the
agent. This algorithm is a BFS because of the behavior of
expanding the search outward from the agent.
%Since BFS is very widely used for applications similar to
%this, and it is outside the scope of this work, the proof of
%this algorithms validity will not be examined here.
The purpose of the solvable gridworld map is to randomly generate
the locations of the entities (agent, obstacles and target).

\begin{figure}
     \centering
     \begin{subfigure}[b]{0.3\textwidth}
         \centering

    \begin{BVerbatim}
-------
|     |
|   # |
|  # !|
| @   |
|   # |
-------
    \end{BVerbatim}
    %\vspace{-1.5em}
         \includegraphics[width=0.8\textwidth]{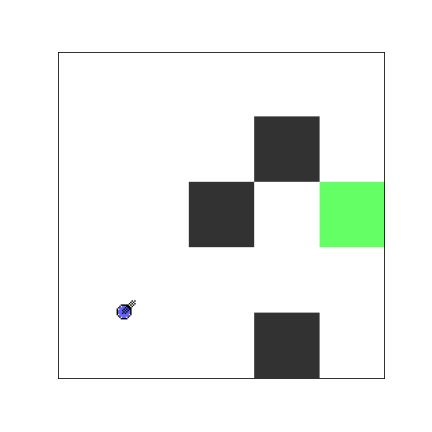}

         \includegraphics[width=\textwidth]{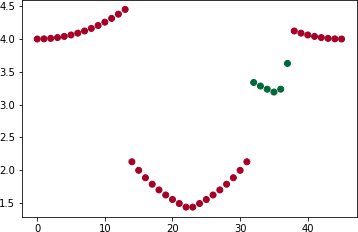}
         \caption{}
         \label{fig:exenv0}
     \end{subfigure}
     \hfill
     \begin{subfigure}[b]{0.3\textwidth}
         \centering
         \begin{BVerbatim}
-------
| !   |
|     |
|  #  |
|     |
|  @  |
-------
         \end{BVerbatim}
    %\vspace{-1.5em}
         \includegraphics[width=0.8\textwidth]{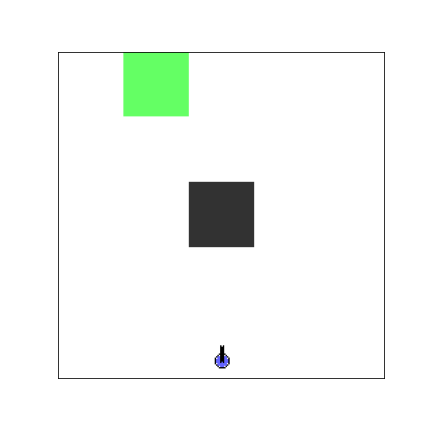}
         
         \includegraphics[width=\textwidth]{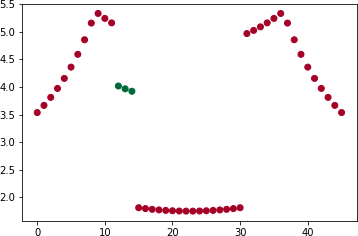}
         \caption{}
         \label{fig:exenv1}
     \end{subfigure}
     \hfill
     \begin{subfigure}[b]{0.3\textwidth}
         \centering
         \begin{BVerbatim}
-------
|   @ |
|     |
|###  |
|     |
|  !  |
-------
         \end{BVerbatim}
         \includegraphics[width=0.8\textwidth]{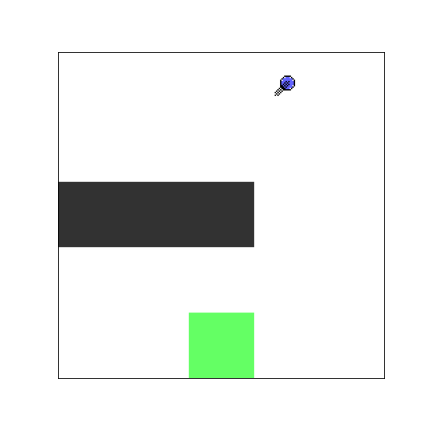}
         
         \includegraphics[width=\textwidth]{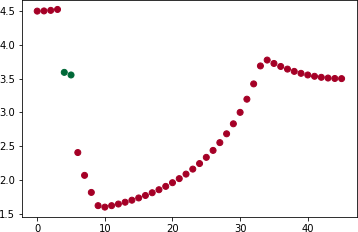}
         \caption{}
         \label{fig:exenv2}
     \end{subfigure}
        \caption{Three example environments with their
        rendered 2D conversions and example agent views}
        \label{fig:exenvs}
\end{figure}

These randomly generated environments and entity positions
are then converted to a 2D environment where the agent can
move by turning or moving forward. This mimics how movement
would occur in a realistic environment. Three examples of
the environments after conversion along with the initial
agent view are shown in Figure~\ref{fig:exenvs}. In this
figure, the target is rendered as a green box, while
obstacles are black.  The blue circle is the the agent. The
direction it is viewing is indicated by the black tick
extending from the blue circle.  The agent is initialized to
a random location within the cell specified by the
``\texttt{@}'' symbol in the gridworld map. The direction it
is viewing is also randomly initialized.

\section{Reinforcement Learning Configuration}
\label{sec:rl}

Using the view angle (the direction towards which the agent
is facing) and location of the agent, a field of view (FOV)
is defined that allows the agent to see the objects in front
of it.  This FOV is divided into slices, and for each slice,
the distance from the agent to the object that the slice
intersects with, along with the type of object, is found
using a 2D ray-tracing algorithm. The type of the object is
encoded as a 0 or a 1, where 1 is a target, and 0 denotes
anything else, be it an obstacle or boundary. This trait is
shown in Figure~\ref{fig:exenvs}. In the bottom row of
graphs, the height of the points from the $x$ axis denotes
the distance from the agent to the object, and the color
denotes the type of the object. The green colored points
correspond to distances to the target, while red is a
distance to a boundary or obstacle. This allows the agent to
recognize that the boundary and obstacles should both be
avoided in the same way.

\begin{wraptable}{r}{5.5cm}
    \centering
    \small
    \vspace{-1.0em}
    \captionof{table}{Reward Function used to train RL models}
    \vspace{-1.0em}
    \begin{tabular}{|c|c|}
        \hline
        \textbf{Condition}     & \textbf{Reward} \\ \hline\hline
        Near Obstacle & -1.0 \\ \hline
        Near Boundary & -1.0 \\ \hline
        Moved Away From Target & -1.5 \\ \hline
        Moved Toward Target & -0.2 \\ \hline
        Reached target & 0.0 \\ \hline
        Default  & -0.7 \\ \hline
    \end{tabular}
    \label{tab:reward}
    \vspace{-1em}
\end{wraptable}

This view of the agent is used as input to the reinforcement
learning models. The agent location, agent view angle, and
the location of the center of the target are included as
input for some models. Models that only receive the visual
input are referred to as ``Pure'' models, whereas those that
receive information on the location of the agent and the
target are referred to as ``Impure.'' This distinction is
made because the pure vision-based models will be required
to seek the targets location in the environment, while the
impure models will mainly need to learn how to avoid
obstacles while approaching a known target.

The reward function used for training the RL models is
manifold. First, if the agent is near an obstacle or
boundary, there is a harsh penalty. Second, if the agent
moves away from the target, the harshest penalty is applied,
but moving toward the target applies a very low penalty.
Reaching the target gives no penalty, and the last case is
kept as default where a medium penalty is applied as a move
cost. The values are shown in Table~\ref{tab:reward}.

This reward function was designed to create a ``punishment
crater'' centered on the target. Approaching the target
applies low penalty, as if the agent is moving down a hill,
but moving away increases the penalty sharply, similar to
ascending a gradient. Being near an obstacle or the boundary
is penalized sharply to encourage the agent to avoid them
while it approaches the target.

The discount factor used is 0.99 for all time steps. When
the agent reaches the target, a discount of 0.0 is given, as
the episode has ended. This encourages the agent to consider
rewards far in the future during decision making so that it
can attempt to reach the goal as quickly and efficiently as
possible during its search.

\begin{figure}[]
     \centering
     \begin{subfigure}[b]{0.19\textwidth}
         \centering
         \includegraphics[width=\textwidth]{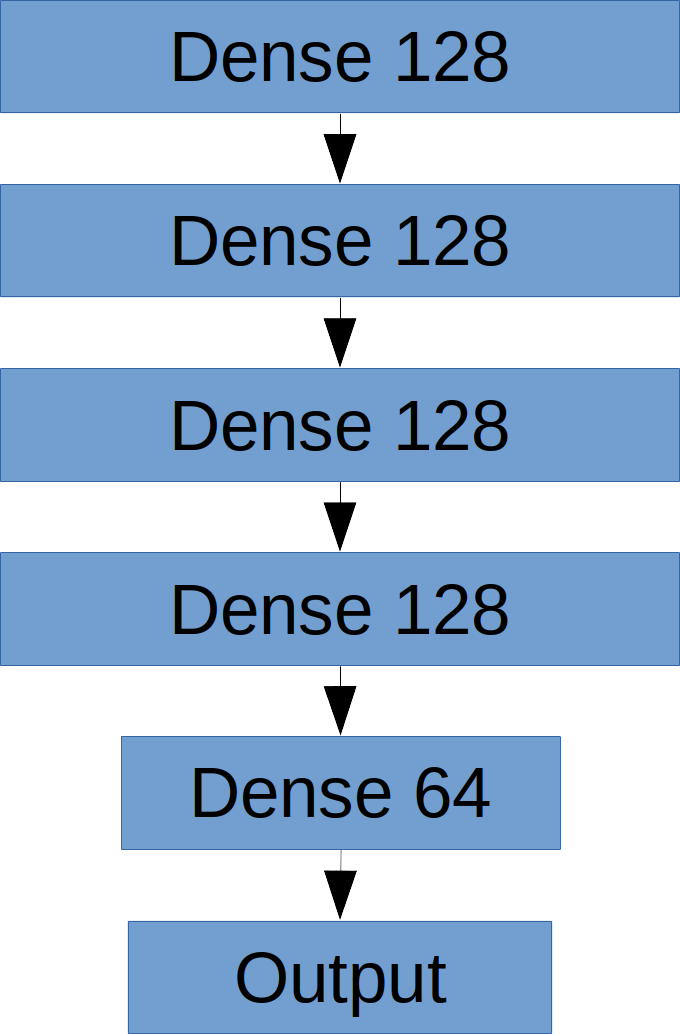}
         \caption{DQN}
         \label{fig:moddqn}
     \end{subfigure}
     \hspace{2em}
     \begin{subfigure}[b]{0.19\textwidth}
         \centering
         \includegraphics[width=\textwidth]{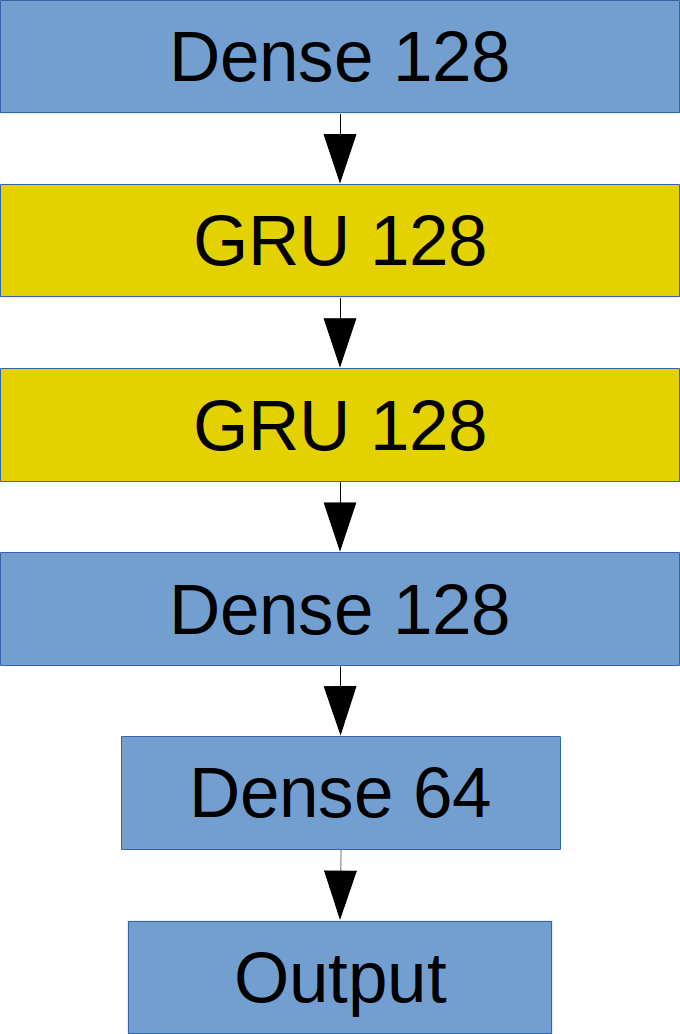}
         \caption{DQN-GRU}
         \label{fig:modgru}
     \end{subfigure}
     \hspace{2em}
     \begin{subfigure}[b]{0.19\textwidth}
         \centering
         \includegraphics[width=\textwidth]{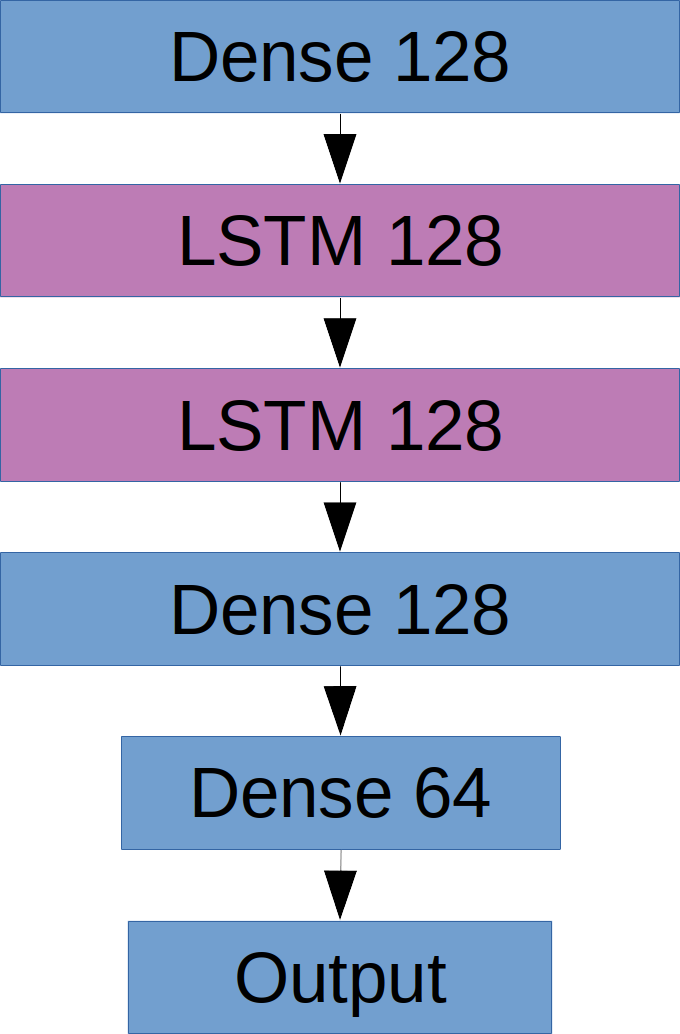}
         \caption{DQN-LSTM}
         \label{fig:modlstm}
     \end{subfigure}
        \caption{Architectures of the Evaluated RL models}
        \label{fig:models}
\end{figure}

In total, six different models are trained and evaluated.
They vary in the type of neural network used to construct
the RL model and the input given to the model. First, a DQN
is used, with only fully connected layers. The second model
replaces the second two layers of the DQN with GRU layers.
This changes the model into a DQRNN using GRU, hereafter
referred to as the DQN-GRU. The third model replaces the GRU
layers with LSTM cells. This model, similar to the second,
is referred to as the DQN-LSTM. Each model uses a very
similar architecture, to promote a fair comparison between
them. Their architectures are shown in
Figure~\ref{fig:models}.  For each model architecture,
models with both the pure and impure versions of input are
trained and evaluated.

\begin{wraptable}{r}{4.5cm}
    \centering
    \small
    \vspace{-1.5em}
    \captionof{table}{Discretized Actions}
    \begin{tabular}{|c||c|c|}
        \hline
        \multirow{2}{*}{\textbf{Discrete}}     & \multicolumn{2}{|c|}{\textbf{Continous}} \\\cline{2-3}
                              & Move & Turn  \\ \hline\hline
        0 &0.05&-1   \\ \hline
        1 &0.5 &-0.5 \\ \hline
        2 & 1  &   0 \\ \hline
        3 &0.5 &0.5  \\ \hline
        4 &0.05& 1   \\ \hline
    \end{tabular}
    \label{tab:acts}
    \vspace{-0.5em}
\end{wraptable}

These three models were chosen as they are all widely-used
DNN and RNN methods. Furthermore, each model is expected to
exhibit a different performance based on the design.  While
the DQN should handle an environment where the target
is visible well, it will likely struggle when obstacles
are introduced. On the other hand, the DQN-LSTM should have
more memory power to handle more complex search
environments. The DQN-GRU has a memory mechanism, but is
less flexible than the DQN-LSTM, so it may not perform as
well.

The Seeker environment receives action input as a pair of
floating point values. The first element of this pair is the
forward/backward movement. A positive value means that the
agent should move forward, while a negative value
corresponds to reverse movement. The magnitude of this value
corresponds to the distance to be moved. The second element
of the pair corresponds to left/right rotation of the agent.
A negative value corresponds to a right turn, while positive
values correspond to left. The magnitude of the value
defines the amount the agent should turn. For this work,
these actions are discretized into five integer numbered
actions. These actions along with their corresponding
move/turn pairs are shown in Table~\ref{tab:acts}. Notice
that reverse movement is disallowed, and all actions include
at least a minimal forward movement component. This
encourages the agent to always move, so that it can be
exposed to the varying reward values, and make optimal
decisions as it moves. Furthermore, in this environment, the
agent does not need to stay in a fixed position. While
rotation without movement may enable more optimal behavior,
in terms of path length, removing this behavior enriches the
variations of the reward function seen by the agent during
the training process. This enrichment of the reward function
enables a more efficient training process by preventing the
model from choosing a behavior where it remains stationary.
Instead, the model is forced to make an action, and
therefore must learn to choose the optimal action.

\begin{figure}[h]
    \centering
    \includegraphics[width=0.9\textwidth]{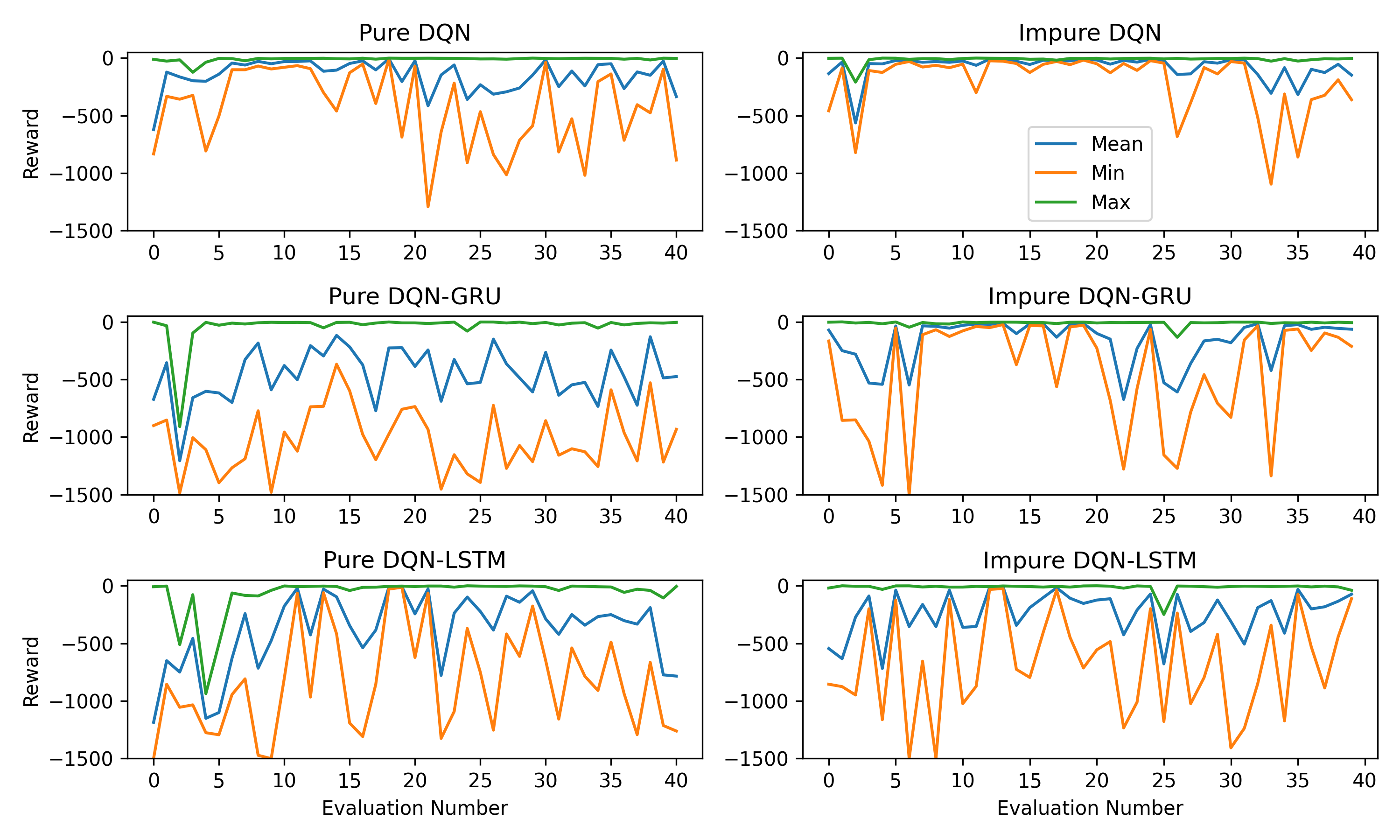}
    \caption{Reward from Intermediate Evaluation During Training}
    \label{fig:trainrd}
\end{figure}

\section{Evaluation}
\label{sec:eval}

\begin{table}
    \centering
    \subcaptionbox{Pure DQN}{
    \begin{tabular}{|c|c|c|c|}
        \hline
        Number of & Episode & \multirow{2}{*}{Reward} & Path   \\
        Obstacles & Length  &                         & Length \\ \hline
           2      & 273.00  & -165.80&4.89         \\ \hline
           4      & 263.45  & -176.21&4.46         \\ \hline
           7      & 453.25  & -288.17&6.66         \\ \hline
    \end{tabular}
    }
    \hspace{0.25in}
    \subcaptionbox{Impure DQN}{
    \begin{tabular}{|c|c|c|c|}
        \hline
        Number of & Episode & \multirow{2}{*}{Reward} & Path   \\
        Obstacles & Length  &                         & Length \\ \hline
           2      &95.62    &-51.20  &3.04         \\ \hline
           4      &117.46   &-45.44  &2.76         \\ \hline
           7      &180.07   &-60.28  &3.35         \\ \hline
    \end{tabular}
    }\\
    \vspace{1em}

    \subcaptionbox{Pure DQN-GRU}{
    \begin{tabular}{|c|c|c|c|}
        \hline
        Number of & Episode & \multirow{2}{*}{Reward} & Path   \\
        Obstacles & Length  &                         & Length \\ \hline
           2      & 394.51  &-340.93 & 5.39        \\ \hline
           4      & 373.63  &-297.76 & 5.01        \\ \hline
           7      & 536.28  &-428.15 & 6.70        \\ \hline
    \end{tabular}
    }
    \hspace{0.25in}
    \subcaptionbox{Impure DQN-GRU}{
    \begin{tabular}{|c|c|c|c|}
        \hline
        Number of & Episode & \multirow{2}{*}{Reward} & Path   \\
        Obstacles & Length  &                         & Length \\ \hline
           2      &149.43   &-87.40  &3.05         \\ \hline
           4      &176.04   &-106.77 &3.16         \\ \hline
           7      &302.66   &-157.27 &3.87         \\ \hline
    \end{tabular}
    }\\
    \vspace{1em}

    \subcaptionbox{Pure DQN-LSTM}{
    \begin{tabular}{|c|c|c|c|}
        \hline
        Number of & Episode & \multirow{2}{*}{Reward} & Path   \\
        Obstacles & Length  &                         & Length \\ \hline
           2      & 596.27  &-435.83 &4.76         \\ \hline
           4      & 527.01  &-384.67 &4.12         \\ \hline
           7      & 693.61  &-567.16 &5.02         \\ \hline
    \end{tabular}
    }
    \hspace{0.25in}
    \subcaptionbox{Impure DQN-LSTM}{
    \begin{tabular}{|c|c|c|c|}
        \hline
        Number of & Episode & \multirow{2}{*}{Reward} & Path   \\
        Obstacles & Length  &                         & Length \\ \hline
           2      &308.71   &-145.68 &3.30         \\ \hline
           4      &318.37   &-147.90 &3.25         \\ \hline
           7      &441.67   &-191.00 &4.03         \\ \hline
    \end{tabular}
    }

    \caption{Average Evaluation Results for All Models}
    \label{tab:eval}
\end{table}

For evaluation, all six models were trained with the same
random sequence of maps for the same number of time steps.
The learning rate used was $1\times10^{-5}$. First, each
model is trained for 500k time steps in a sequence of maps
without any obstacles; only the map boundary and target are
present. Then, 3 obstacles are added before training for
another 250k time steps. Last, a total of 5 obstacles are
used for 250k time steps of training. This totals to 1
million time steps of training for each model.  The number
of episodes the model is exposed to is at least 1000, as the
maximum length of an episode is 1000 time steps. However,
due to the episodes being able to end early, the models may
be exposed to more episodes depending on their learning
capabilities. These increases in the number of obstacles
present in the environment map allow for the agent to learn
an easier version of the problem before it attempts to learn
a more difficult version. This is done with the goal of
training the models more quickly than if training was only
performed in a difficult environment.

Each number of obstacles used during training defines a
phase of training, meaning there are in total 3 phases of
training. At every 25k time steps of training, training is
halted to evaluate the models performance on a random set of
5 maps. So, for the first training stage, there are 20
evaluations, and for the second and third, 10 evaluations
occur. This gives a view of how well the model has learned
how to search for the target at that point in training.
Figure~\ref{fig:trainrd} shows the results of this
evaluation. Each row of the figure corresponds to a model
architecture, while each column corresponds to the type of
input used. The DQN models perform well when no obstacles
are present, but cannot maintain their performance
consistently as more obstacles are added. The Impure DQN-GRU
model performs fairly well, and is able to learn how to
handle obstacles much better than the DQNs were able to.
Lastly, the DQN-LSTM model is shown. The impure version of
model does fairly well, in terms of reward, but throughout
training it is less stable than the DQN-GRU model, even if
the average does tend to improve.

To thoroughly compare the capabilities of each of the three
models, a set of tests are performed. Each model is tested
using 2, 4, and 7 obstacles. These values are similar to the
number of obstacles that are used during training, but 7 is
higher than any that the agents see during training. This
set of obstacle numbers allows for a comparison of the model
performances when they are in environments similar to those
used for training, along with a more difficult environment
to examine each models generalization ability.

For each of the three chosen numbers of obstacles, each
agent is evaluated in a sequence of 100 random environment
maps. This sequence is the same for each agent. The average
reward per episode, average episode length in time steps,
and average agent path length are shown in
Table~\ref{tab:eval}.  This evaluation shows that the DQN
model is able to learn the environment much faster than the
RNN versions, as the DQN tends to outperform the other
models, especially in the impure case. Interestingly,
however, the pure DQN-LSTM outperforms the pure DQN model in
terms of the path length.  The amount of time taken by the
DQN-LSTM tends to be much higher than the DQN, and the
reward is much lower, but it shows that it is able to find
shorter paths to the target.  This is likely due to the
LSTMs capability to make better use of previously seen
information when making decisions.  The pure DQN-GRU model
also has a memory capability, but it finds longer paths than
either the DQN or DQN-LSTM model, and the pure DQN
outperforms it in terms of episode length and reward.

\section{Conclusion and Future Work}
\label{sec:conc}

% Conclusion

Pathfinding is a key problem for enabling autonomous robots,
drones, and vehicles. Reinforcement learning provides a
method to train models to perform pathfinding in a variety
of environments, but many current research works focus on
gridworld environments. These environments restrict the
capability of the trained models, as any real world
application of these models would need to rely on
translation of the real environment into a grid. In this
work, a simulated environment for training RL models named
Seeker is introduced. The Seeker environment enables the
training of models based on visual input to make realistic
movement decisions. By informing models with vision, they
are better prepared to be deployed in real world
applications.

The evaluation performed in this work demonstrates the
capabilities of three popular RL models in the Seeker
environment, namely DQN, DQN-GRU, and DQN-LSTM. Each of
these models utilizes a different mechanism for making
decisions, and two have memory capabilities. Interestingly,
the DQN model outperformed its recurrent counterparts when
trained in an equivalent manner. This points to a need for
differing training regimens for each model to obtain optimal
performance.

% Future Work

Expansion of this work may include the addition of
convolutional layers to the neural network models used. This
addition will enable the extraction of learned spatial
features from the vision, and potentially benefit the RL
models' capability to make decisions. Beyond this, there are
many other RL models whose performance in this environment
can be evaluated, both with discrete action sets and using
the native continuous movement scheme. Furthermore, a
thorough investigation into the recurrent models performance
should be performed to determine the models optimal ability
level in the Seeker environment.

{\footnotesize
\bibliographystyle{ieeetr}
\bibliography{ms}
}
\end{document}